\documentclass{article}
\usepackage{arxiv}

\usepackage[T1]{fontenc}
\usepackage[utf8]{inputenc}
\usepackage{amsmath,amssymb}
\usepackage{graphicx}
\usepackage{booktabs,multirow}
\usepackage{microtype}
\usepackage{placeins}
\usepackage{etoolbox}
\usepackage[numbers,sort&compress]{natbib}
\usepackage{xurl}
\usepackage[hidelinks]{hyperref}
\usepackage[font=small,labelfont=bf]{caption}
\hypersetup{
  pdftitle={CoptV2X: Coupling-Aware Bilevel Optimization for V2X Perception Model Compression and Backend Scheduling},
  pdfauthor={Chengzhi Ji, Yuming Ge, Ziyuan Pu, Cheng-Zhong Xu},
  pdfsubject={Cooperative perception and software--hardware co-optimization},
  pdfkeywords={V2X, cooperative perception, model compression, backend scheduling, bilevel optimization}
}

\newcommand{\bestcell}[1]{\textbf{#1}}
\newcommand{\secondcell}[1]{\underline{#1}}
\title{CoptV2X: Coupling-Aware Bilevel Optimization for V2X Perception Model Compression and Backend Scheduling}
\author{%
  \begin{minipage}{\textwidth}
  \centering
  \normalfont\normalsize
  Chengzhi Ji\textsuperscript{1}\quad
  Yuming Ge\textsuperscript{2}\quad
  ChengZhong Xu\textsuperscript{3}
  Ziyuan Pu\textsuperscript{1,*}\quad
  \\[8pt]
  \small
  \textsuperscript{1}School of Transportation, Southeast University, Nanjing 211189, China\\[5pt]
  \textsuperscript{2}Key Laboratory of Internet of Vehicle Technical Innovation and Testing, Ministry of Industry and Information Technology,\\
  China Academy of Information and Communications Technology, Beijing 100191, China\\[5pt]
  \textsuperscript{3}Department of Computer and Information Science, Faculty of Science and Technology, University of Macau, Macao SAR, China\\[7pt]
  \textsuperscript{*}Corresponding author:
  \href{mailto:ziyuanpu@seu.edu.cn}{\texttt{ziyuanpu@seu.edu.cn}}
  \end{minipage}%
}
\makeatletter
\patchcmd{\@maketitle}{\vskip 0.4in \@minus 0.1in \center{\today}   \vskip 0.2in}{\vskip 0.15in}{}{%
  \PackageError{CoptV2X}{Could not suppress template date}{Check arxiv.sty.}}
\makeatother
\date{}

\begin{document}
\raggedbottom
\maketitle

\begin{abstract}
Cooperative vehicle-to-everything (V2X) perception extends sensing coverage, but deploying its computationally demanding models on resource-constrained edge devices requires balancing accuracy, latency, and energy consumption.
Model compression and backend scheduling are often optimized separately, even though pruning and quantization reshape operators and execution paths, while backend capabilities determine whether those changes yield deployment gains.
We present CoptV2X, a coupling-aware bilevel optimization framework that coordinates hardware-aware compression search with candidate-specific execution-schedule optimization.
Computation-graph dependencies and hardware capabilities jointly define a compact feasible search space.
Within this space, an outer loop explores pruning and quantization configurations, an inner loop tunes backend execution, and multi-metric cost models prioritize promising candidates under a limited hardware-measurement budget.
Measured results feed back into subsequent search rounds.
We evaluate three representative cooperative-perception models on DAIR-V2X and OPV2V, using 1,789 and 2,170 evaluation samples, respectively, across NVIDIA H800 and Jetson AGX Orin platforms.
Under an accuracy-loss budget of $\Delta\mathrm{AP}_{70}\leq0.10$, CoptV2X reduces latency and energy consumption by 28.03\% and 32.94\% on average, respectively, over the strongest feasible decoupled baseline across the two H800 backend profiles, while achieving an average inference speedup of $12.94\times$ over the original deployments on Jetson AGX Orin.
These results demonstrate the value of jointly optimizing model compression and backend execution for efficient V2X perception deployment.
\end{abstract}

\section{Introduction}

\begin{figure}[!htbp]
\centering
\includegraphics[width=\columnwidth]{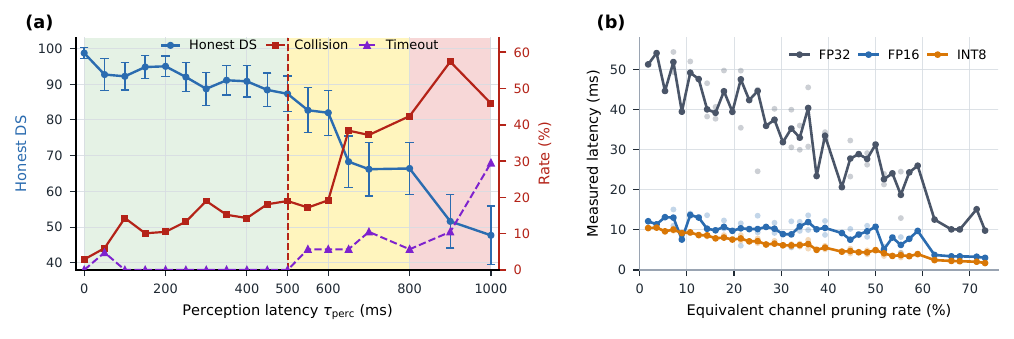}
\caption{Motivating evidence for software--hardware co-optimization. (a)
Closed-loop performance under injected perception latency. (b)
Measured latency of channel-pruned candidates under FP32, FP16, and INT8 execution.}
\label{fig:bg-latency-prune}
\end{figure}

Vehicle-to-everything (V2X) cooperative perception has become an important paradigm for safer and smarter autonomous driving.
By connecting roadside units (RSUs) with onboard sensors, it builds a shared perception system that extends each vehicle’s observable range and improves scene understanding under occlusion~\cite{yu2025univ2x,wang2026sparsecoop}. 
Since V2X perception depends on timely information exchange and fusion across agents, delayed messages may inject stale observations into fused perception.
This can cause a mismatch with the current traffic state and potentially affect downstream planning and control.
As shown in Fig.~\ref{fig:bg-latency-prune}(a), by injecting perception latency into the V2XVerse closed-loop driving evaluation~\cite{liu2025toward}, we observe that the driving score (DS) decreases as perception latency increases, while collision and timeout rates rise.
These observations demonstrate that cooperative perception must operate within a limited latency budget to preserve closed-loop driving performance. 
Meanwhile, the complex feature-extraction, fusion, and detection modules used by current V2X models impose substantial latency and energy costs on edge platforms. 
Efficient V2X perception deployment must therefore reduce both inference latency and energy consumption without materially degrading perception accuracy~\cite{liu2026ipda,wang2026cpfreezer}.

From the perspective of the Roofline model~\cite{williams2009roofline}, inference latency can be reduced through two complementary paths~\cite{shen2022halp}: reducing the workload through pruning and quantization, and improving execution efficiency through compilation, operator scheduling, and memory-access optimization.
In practice, these two paths are tightly coupled~\cite{chen2021chanas,xu2023alt}. 
Compression changes operator shapes and data formats, thereby inducing different feasible and efficient backend schedules.
Conversely, hardware capabilities and backend support determine whether compression yields actual deployment acceleration.
The measurements in Fig.~\ref{fig:bg-latency-prune}(b) support this point: across 60 channel-width candidates under FP32, FP16, and INT8 execution, latency generally decreases with stronger pruning.
However, the curves exhibit crossings and fluctuations, and the acceleration benefit of quantization varies with channel configuration. 

Recent studies have made substantial progress in both software compression search and hardware execution optimization. 
On the software side, methods assign mixed precision bit-widths to different layers~\cite{yang2021fracbits}, search for joint pruning-quantization configurations~\cite{hu2021opq,balasubramaniam2025upaq} and further use automated search to identify better compression configurations~\cite{xiao2023haloc}. 
On the hardware side, methods generate efficient kernels through fast tensor compilation~\cite{zhu2022roller}, build shared scheduling spaces for dynamic shapes~\cite{zheng2022dietcode} and use tensor-program search to generate better execution configurations~\cite{shao2022metaschedule}.
However, existing methods still commonly model the software and hardware optimization spaces separately. 
This separation leads to two limitations: compression methods often lack feedback from hardware execution constraints, while compilation and scheduling methods usually optimize a fixed computational graph and provide limited backward guidance for model compression. 
Consequently, the optimization process may easily converge to suboptimal solutions.
Some recent studies have explored joint optimization of model structure and execution strategy, such as architecture--schedule co-search~\cite{chen2021chanas} and graph-layout/loop-level co-optimization~\cite{xu2023alt}. However, they still provide limited characterization of the bidirectional dependence between pruning/quantization configurations and the hardware scheduling space they induce.

To address these challenges, we propose CoptV2X, a bilevel multi-objective software--hardware co-search method for efficient V2X perception deployment.
CoptV2X focuses on two core problems: how to construct a compact, scenario-specific feasible space under graph and backend constraints and how to efficiently explore this space for Pareto-optimal deployments.
Our contributions are summarized as follows:

\begin{itemize}
\item We formulate V2X perception deployment as constrained multi-objective optimization over structured pruning, numerical precision, and backend schedules. 
We develop a dual-constraint space-construction method that derives a compact, scenario-specific feasible space under graph-dependency and hardware-capability constraints.

\item We construct a bilevel framework in which the outer loop explores compression configurations with hardware feedback, while the inner loop optimizes candidate-specific execution schedules.
This structure preserves software--backend coupling without flattening the joint space, while multi-metric cost models screen candidates before inner-loop tuning to reduce measurement cost.

\item We evaluate CoptV2X on representative V2X perception models across multiple deployment backends and hardware platforms.
Under a common accuracy-loss budget, CoptV2X reduces latency and energy consumption by 28.03\% and 32.94\% on average, respectively, over the strongest feasible decoupled baseline on H800 with only marginal accuracy degradation, while delivering an average inference speedup of $12.94\times$ over the original deployments on Jetson AGX Orin.

\end{itemize}

\FloatBarrier
\section{Method}

\subsection{Overview}

CoptV2X is a software--hardware co-optimization framework for efficient deployment of cooperative V2X perception models on edge devices. 
It automatically searches for deployment configurations that balance perception accuracy, inference latency, and energy consumption. 

\begin{figure}[!htbp]
    \centering
    \includegraphics[width=0.98\textwidth]{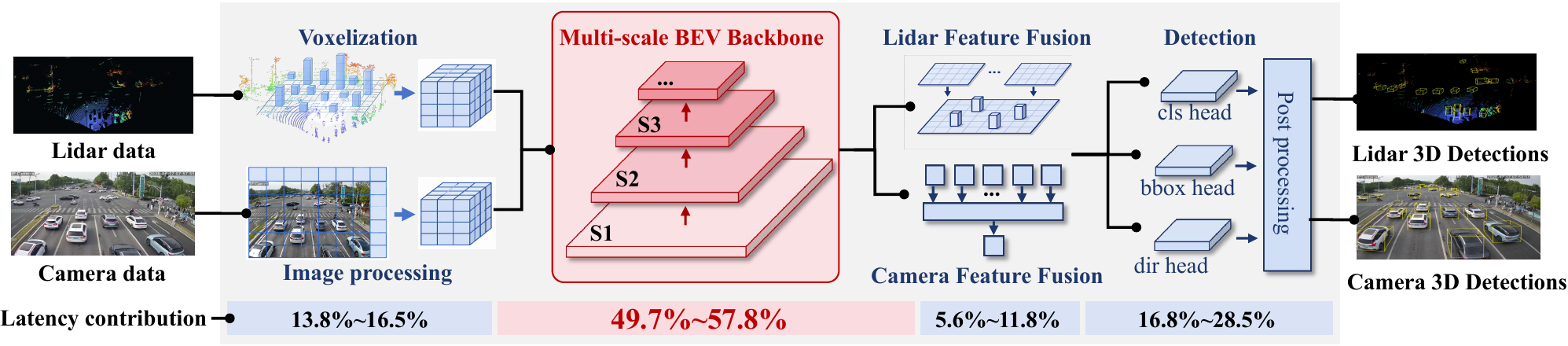}
    \caption{Module-wise latency breakdown of representative V2X perception models under the target hardware and backend.}
    \label{fig:latency-contribution}
\end{figure}

Fig.~\ref{fig:latency-contribution} presents the module-wise contributions to the end-to-end inference latency of representative V2X perception models under the specified hardware and backend configuration. 
The multi-scale bird's-eye-view (BEV) backbone contributes substantially more to the inference latency than the other modules and constitutes a major computational bottleneck on both the RSU and vehicle sides. 
CoptV2X therefore treats the multi-scale BEV backbone as the primary target for structural optimization.

\begin{figure}[!htbp]
\centering
\includegraphics[width=0.58\linewidth]{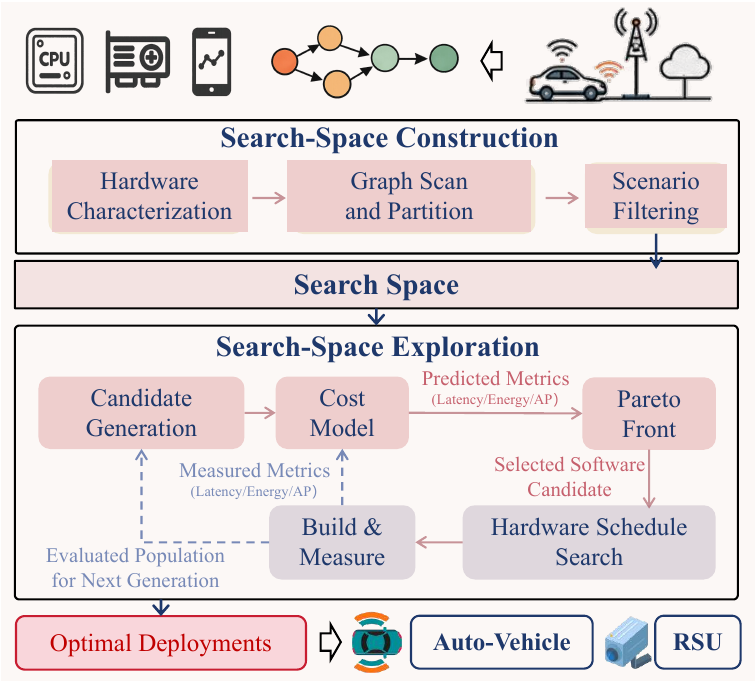}
\caption{Overview of CoptV2X. The framework constructs a scenario-specific search space from the target model and hardware platform, then explores pruning, quantization, and scheduling configurations with measurement feedback to obtain Pareto deployment candidates.}
\label{fig:framework}
\end{figure}

As shown in Fig.~\ref{fig:framework}, CoptV2X consists of two major modules: search-space construction and search-space exploration. 
The search-space construction module defines a general procedure for deriving a hardware-aware design space from a given model and target device. 
Based on this space, the search-space exploration module performs efficient multi-objective co-search across pruning, quantization, and scheduling decisions, and outputs Pareto deployment candidates for different deployment requirements.

\FloatBarrier
\subsection{Dual-Constraint Search-Space Construction}

\begin{figure}[!htbp]
\centering
\includegraphics[width=0.86\linewidth]{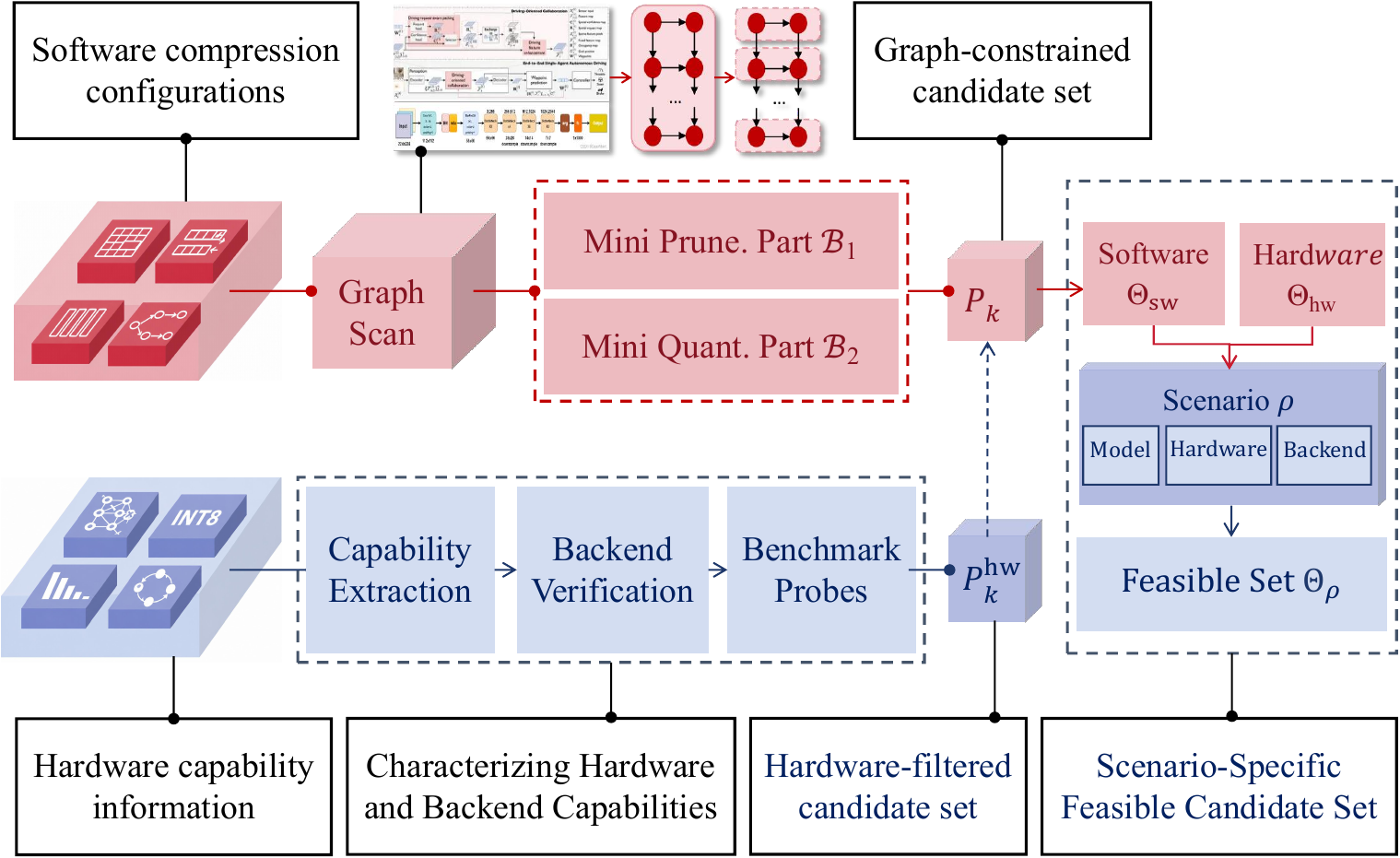}
\caption{Search-space construction in CoptV2X. Graph dependency partitioning and hardware capability characterization jointly filter the raw search spaces to form a scenario-specific feasible set.}
\label{fig:search-space-construction}
\end{figure}

The joint design space consists of software compression configurations and hardware scheduling configurations, as illustrated in Fig.~\ref{fig:search-space-construction}.

In the software compression dimension, structured pruning and post-training quantization are used as the main optimization knobs. 
Compared with training-intensive compression alternatives, these two choices reduce computation and storage cost while requiring less retraining and architectural redesign~\cite{zhao2019variational,lee2023flexround}. 

A software configuration is defined as $x=(p,q)$, where $p$ denotes the channel-width configuration of the dependency-coupled structural groups, and $q$ denotes the quantization bit width.
All software configurations that satisfy hardware capability constraints and computation-graph structural constraints form the software search space $\Theta_{\mathrm{sw}}$.

In the hardware scheduling dimension, we follow the scheduling-space definitions used in existing tensor-program auto-tuning methods~\cite{chen2018learning,shao2022metaschedule}.
The hardware scheduling space, denoted by $\Theta_{\mathrm{hw}}$, consists of backend optimization configurations involving loop tiling, memory layout, vectorization, and other schedule-level transformations. A hardware scheduling configuration is denoted by $s$. 
A complete deployment candidate is denoted by $c=(x,s)$.

For a given deployment scenario,  we formulate the joint design problem as the constrained multi-objective optimization task in Equation~\ref{eq:pareto-objective}:
\begin{equation}
\mathcal{P}^{\star}_{\rho} =
\mathrm{Pareto}_{c\in\Theta_{\rho}}
\bigl(-\mathrm{AP}(c), \ell(c), e(c)\bigr).
\label{eq:pareto-objective}
\end{equation}
The feasible candidate set for scenario $\rho$ is defined in Equation~\ref{eq:feasible-candidate-set}:
\begin{equation}
\Theta_{\rho}=
\{c=(x,s)\mid x\in\Theta_{\mathrm{sw}},s\in\Theta_{\mathrm{hw}},
g_{\rho}(c)\in\mathcal{C}_{\rho}\}.
\label{eq:feasible-candidate-set}
\end{equation}
Here, $\rho$ denotes a deployment scenario specified by the target model, target hardware platform, and backend capability profile. 
$\mathrm{AP}(c)$, $\ell(c)$, and $e(c)$ denote the detection accuracy, latency, and energy consumption of candidate $c$ on the target hardware, respectively.
$\mathcal{C}_{\rho}$ is the feasible region defined by hardware capability and deployment budget constraints, and $g_{\rho}(c)$ maps candidate $c$ to the corresponding constraint attributes.
$\mathcal{P}^{\star}_{\rho}$ is the set of Pareto-optimal candidates that satisfy the scenario constraints.

The software compression space is already prohibitively large due to fine-grained pruning choices and the complex architecture of cooperative V2X perception models.
Coupling quantization bit width with hardware scheduling further expands the raw joint space, making direct search impractical.
We therefore explicitly construct software and hardware constraints through hardware capability characterization and computation-graph dependency partitioning, thereby compressing the raw space into a tractable set of configurations for subsequent optimization.

\subsubsection{Hardware Capability Characterization}

We describe the target hardware capability by $\mathcal{H}$ through three steps.
(1) we extract the execution capabilities of the target hardware and compiler backend, covering operator support, numerical precision, quantization granularity, and memory-access constraints.
(2) we compile candidate operators and subgraphs to verify whether they can be accepted by the target backend.
(3) we run a small number of micro-benchmark probes to identify discrete constraints such as channel packing and low-precision kernel trigger conditions.

Based on $\mathcal{H}$, raw candidates are converted into deployment-relevant discrete configurations.
For the $k$-th upper-level partition, the hardware-feasible local software candidate set is defined in Equation~\ref{eq:hardware-filter}:
\begin{equation}
P_k^{\mathrm{hw}} =
\{x_k\mid x_k\in\Omega_k,\ \psi_{\mathrm{hw}}(x_k,\mathcal{H})=1\}.
\label{eq:hardware-filter}
\end{equation}
Here, $P_k^{\mathrm{hw}}$ is the local software candidate set retained after hardware capability filtering; $\Omega_k$ is the pre-constraint candidate set of the partition, including pruning and quantization choices; and $\psi_{\mathrm{hw}}$ is the deployability predicate induced by $\mathcal{H}$, which excludes software configurations unsupported by the hardware or backend.

\subsubsection{Computation-Graph Scanning and Partitioning}

To preserve graph consistency and tensor-shape legality during pruning, we adopt dependency-aware structured pruning based on computational graphs~\cite{fang2023depgraph}. 
The network graph is scanned to group channel-coupled operators in dense and shape-fixed computation cores into dependency blocks. 
Each block defines a minimal pruning partition in $\mathcal{B}_1$, and pruning knobs are only assigned to these partitions rather than arbitrary individual layers. 
Quantization partitions $\mathcal{B}_2$ are constructed by merging complete pruning partitions, avoiding broken channel dependencies at quantization boundaries. 
The resulting graph partition is represented as $\mathcal{G}=\langle\mathcal{B}_1,\mathcal{B}_2\rangle$.

Directly using all low-level dependency blocks as search variables still produces an excessively large space, on the order of $10^{30}$.
Therefore, we further merge dependency blocks with similar semantic positions and identical hardware constraints into upper-level search partitions $\mathcal{G}'$.
For the $k$-th upper-level search partition, Equation~\ref{eq:software-filter} defines the candidate set retained after software-structure filtering:
\begin{equation}
P_k =
\{x_k\in P_k^{\mathrm{hw}}\mid \psi_{\mathrm{sw}}(x_k,\mathcal{G}')=1\}.
\label{eq:software-filter}
\end{equation}
Here, $P_k$ is the local candidate set that satisfies both hardware capability and graph-partition constraints; $\psi_{\mathrm{sw}}$ is the software structure predicate induced by $\mathcal{G}'$, which filters out configurations that break pruning partitions, quantization boundaries, or cross-layer channel dependencies.
The resulting software search space is formed by all upper-level partition candidates, as shown in Equation~\ref{eq:software-space}, where $K$ is the number of upper-level search partitions:
\begin{equation}
\Theta_{\mathrm{sw}}=\prod_{k=1}^{K}P_k.
\label{eq:software-space}
\end{equation}
This gives an enumerable software search space whose candidates already satisfy the hardware capability and graph-structure constraints.

\FloatBarrier
\subsection{Bidirectionally Coupled Search-Space Exploration}
\label{subsec:exploration}

\subsubsection{Bilevel Search Architecture}

\begin{figure}[!htbp]
\centering
\includegraphics[width=0.96\linewidth]{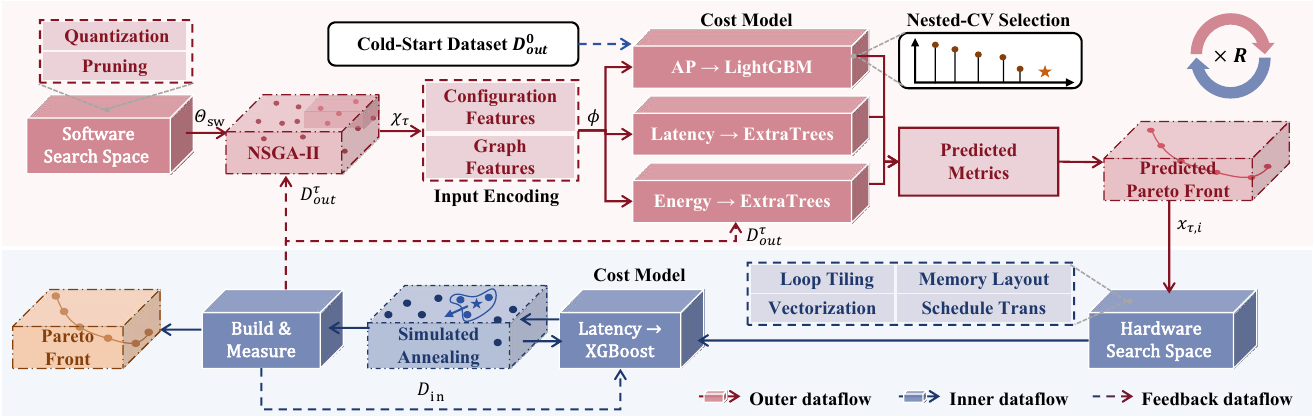}
\caption{Bilevel exploration in CoptV2X. The outer loop searches pruning and quantization configurations, whereas the inner loop optimizes backend schedules and feeds measured results back to the outer loop.
$\mathcal{D}_{\mathrm{out}}^{0}$, $\mathcal{D}_{\mathrm{out}}^{\tau}$, and $\mathcal{D}_{\mathrm{in}}$ denote the cold-start set, accumulated outer-loop observations, and inner-loop tuning records, respectively; $\times R$ denotes $R$ outer iterations. All other symbols are defined in the Method section.}
\label{fig:bilevel-exploration-architecture}
\end{figure}

The main challenge of software--hardware co-search comes from two sources.
First, pruning and quantization configurations $x$ change operator shapes and memory access patterns, making hardware scheduling $s$ conditional on the software configuration.
Second, $\Theta_{\mathrm{hw}}$ has a large number of knobs.
Flattening $x$ and $s$ into a single space would sharply increase dimensionality and generate many invalid candidates.
To preserve this conditional dependence while controlling search dimension, we construct the bilevel exploration architecture illustrated in Fig.~\ref{fig:bilevel-exploration-architecture}.

Specifically, we develop a surrogate-assisted multi-objective evolutionary optimization framework for the outer search. Building on the sequential surrogate-update principle of sequential model-based optimization (SMBO)~\cite{lindauer2022smac3,rook2026mosmac}, our framework incorporates multi-objective evolutionary candidate generation, predicted-Pareto-frontier screening, and conditional hardware-scheduling feedback, thereby avoiding a flattened joint search over software configurations and low-level schedules.

It uses the pruning and quantization configuration $x$ as the explicit search variable, and obtains hardware response signals through scheduling probes.
At the beginning of search, the outer cost model is initialized using a cold-start dataset constructed from measured candidates and their multi-metric observations.
Then, the multi-objective evolutionary strategy generates candidate configurations in $\Theta_{\mathrm{sw}}$, and the cost model rapidly evaluates them.
Only candidates near the predicted Pareto frontier enter inner scheduling.

The best schedule found by the inner search is instantiated and measured on the target hardware.
The resulting metrics are added to the outer observation set to update the cost models and guide subsequent candidate ranking.
This feedback incorporates backend behavior into the outer loop without exhaustively searching low-level schedules.
Pruning and quantization configurations are therefore iteratively re-ranked according to their measured hardware performance.

The inner scheduling search adopts an AutoTVM-style automatic tuning strategy~\cite{chen2018learning}: it generates schedule candidates through simulated annealing and predicts their performance using an XGBoost cost model~\cite{chen2016xgboost}.
A small number of target-hardware measurements are used to calibrate the predictor and select a better schedule implementation.
In addition, the framework provides a unified backend interface that allows other backends, such as TensorRT, to be integrated using their native tuning mechanisms.

\FloatBarrier
\subsubsection{Multi-Metric Cost Models}

Compared with latency-oriented single-objective search in conventional neural architecture search (NAS), multi-metric joint search faces expensive sample acquisition, coupled objectives, and non-smooth metric responses.
Smooth regressors may struggle with such locally irregular responses, whereas multilayer perceptron (MLP) or graph convolutional network (GCN)-based predictors usually require larger training sets~\cite{grinsztajn2022tree,tang2020semi}.

We therefore employ tree-based ensembles to construct target-specific cost predictors for the outer search.
These models capture nonlinear interactions and threshold effects in mixed configuration features, making them well suited to our limited-sample and locally non-smooth evaluation setting.

For the $i$-th candidate $x_{\tau,i}$ at search iteration $\tau$, the predicted value of metric $r$ is defined in Equation~\ref{eq:cost_model_prediction}:
\begin{equation}
\widehat{f}_{r,\tau}(x_{\tau,i})=\mathcal{E}_{r}^{(\tau)}\left(\phi(x_{\tau,i})\right),\quad r\in\{\ell,e,\mathrm{AP}\},
\label{eq:cost_model_prediction}
\end{equation}
where $\mathcal{E}_{r}^{(\tau)}$ denotes the predictor independently trained for metric $r$ at iteration $\tau$. 
The predictors share the same input representation $\phi(x_{\tau,i})$ but use separate model parameters and learning objectives, allowing them to accommodate the distinct scales and measurement characteristics of detection accuracy, latency, and energy consumption.

The mapping from a compression configuration $x$ to its instantiated computational graph is model-dependent and nonlinear. 
Similar compression configurations applied to different perception models may result in substantially different computational and memory characteristics. 
When only configuration variables are provided, a predictor must infer these structural differences from limited observations, which increases the difficulty of cross-model performance prediction. 
To address this limitation, we introduce a graph-aware input representation that combines configuration features with computational graph statistics, as defined in Equation~\ref{eq:graph_aware_input}:
\begin{equation}
\phi(x_{\tau,i})=\left[\phi_{\mathrm{cfg}}(x_{\tau,i}),\,\phi_{\mathrm{graph}}(x_{\tau,i})\right],
\label{eq:graph_aware_input}
\end{equation}
where $\phi_{\mathrm{cfg}}(x_{\tau,i})$ is a numerical characteristic vector that encodes the pruning and quantization decisions associated with the candidate $x_{\tau,i}$ and serves as the basic input to the cost model. 
$\phi_{\mathrm{graph}}(x_{\tau,i})$ summarizes the structure of the computational graph instantiated under the corresponding compression configuration, including the operator composition and operator counts, computational cost, and parameter count.
For each pool candidate, graph features are extracted before prediction through lightweight graph materialization and static analysis, without backend compilation or hardware execution. After deployment, only measured metrics are added to the outer observation set.
This representation enables the predictors to learn nonlinear relationships between the realized graph structure and detection accuracy, latency, and energy consumption.

Latency and energy use independent ExtraTrees regressors~\cite{geurts2006extremely} trained on \texttt{log1p}-transformed labels. 
For $\mathrm{AP}$, we use a LightGBM regressor~\cite{ke2017lightgbm} with the Huber loss and a residual target relative to each model's median training-fold $\mathrm{AP}$.
All predictions are mapped back to their original metric spaces; Appendix~\ref{app:cost-model-selection} details the joint selection of model class, learning objective, and target transformation.

\subsubsection{Candidate Generation}

Candidate generation must preserve diverse trade-offs among $\mathrm{AP}$, latency and energy, rather than compressing them into a fixed scalar.
We therefore instantiate NSGA-II~\cite{deb2002nsga} as the outer candidate generation method.
NSGA-II performs non-dominated sorting and crowding-distance selection based on the predicted metrics, and generates the next round of software candidates from $\Theta_{\mathrm{sw}}$.
This mechanism avoids pre-defining fixed metric weights and maintains Pareto candidates that cover different deployment preferences.

\FloatBarrier

\section{Experiments}

\subsection{Experimental Setup}

\begin{sloppypar}
We evaluate CoptV2X on multiple cooperative perception models, including Pyramid \cite{lu2024heal}, CoDriving \cite{liu2025toward}, and F-Cooper \cite{chen2019fcooper}, to assess its generalizability. 
We use the DAIR-V2X \cite{yu2022dairv2x} and OPV2V \cite{xu2022opv2v} datasets according to each
model's evaluation protocol, with 1,789 and 2,170 evaluation samples, respectively. 
Experiments cover NVIDIA H800 and Jetson AGX Orin hardware platforms, with TensorRT as an alternative backend.
Considering the deployment requirements of V2X perception, we use inference latency, energy consumption, and average precision at an intersection over union (IoU) threshold of 0.70 ($\mathrm{AP}_{70}$) as the primary evaluation metrics.
For each model--backend profile, we enumerate its feasible software candidate pool and run 4 outer iterations with a batch size of 4, resulting in 16 new real-hardware evaluations.
Inner tuning terminates according to backend-specific stopping criteria.
To isolate structural and backend effects, latency is measured as CUDA execution time of the frozen deployed subnet, excluding
data loading and host--device transfers. 
H800 energy is derived from NVML board power, whereas Orin energy uses the active-window \texttt{VIN\_\allowbreak SYS\_\allowbreak 5V0} rail.
Because these measurements have different electrical boundaries, we do not report cross-hardware energy ratios.
\end{sloppypar}

\subsection{Performance Across Hardware Platforms and Backends}
Existing software--hardware co-optimization methods jointly explore network architectures and execution schedules~\cite{chen2021chanas,xu2023alt}. 
However, their optimization objectives, search spaces, and evaluation metrics differ from those used in this work, complicating direct comparisons. 
To enable controlled and reproducible evaluation, we construct four baselines that reflect representative model-deployment pipelines. 
\textit{Original/Default} directly deploys the original model without model compression or scheduling optimization. 
\textit{Compression Only} applies compression techniques, including pruning and quantization, without backend scheduling optimization~\cite{hu2021opq,wang2019haq}. 
\textit{Schedule Only} preserves the original computation graph and searches for efficient implementations exclusively through backend auto-tuning~\cite{chen2018learning,shao2022metaschedule}. 
\textit{Compress} $\rightarrow$ \textit{Tune} represents the sequential workflow commonly used in model deployment. In this workflow, model compression precedes backend-specific scheduling optimization~\cite{zhao2025quantv2x,balasubramaniam2025upaq}.

Given their different objectives and output formats, we adopt a unified representative-point comparison protocol. 
Under the constraint $\Delta\mathrm{AP}_{70}=\mathrm{AP}_{70}^{\mathrm{orig}}-\mathrm{AP}_{70}\leq0.10$, we select the lowest-latency feasible output from each method.
We then compare $\mathrm{AP}_{70}$, latency, and energy consumption under identical hardware, backend, and measurement boundaries. 
The results are reported in Table~\ref{tab:main-comparison}.

\begin{table}[!t]
\centering
\fontsize{8}{10}\selectfont
\setlength{\tabcolsep}{0.75pt}
\begin{tabular*}{\textwidth}{@{\extracolsep{\fill}}l@{\hspace{4pt}}lrrrrrrrrr@{}}
\toprule
& & \multicolumn{3}{c}{Pyramid} & \multicolumn{3}{c}{CoDriving} & \multicolumn{3}{c}{F-Cooper} \\
\cmidrule(lr){3-5}\cmidrule(lr){6-8}\cmidrule(lr){9-11}
Profile & Method
& $\mathrm{AP}_{70}\uparrow$ & Lat. (ms)$\downarrow$ & Energy (J)$\downarrow$
& $\mathrm{AP}_{70}\uparrow$ & Lat. (ms)$\downarrow$ & Energy (J)$\downarrow$
& $\mathrm{AP}_{70}\uparrow$ & Lat. (ms)$\downarrow$ & Energy (J)$\downarrow$ \\
\midrule
\multirow{5}{*}{(a)}
& Original/Default
& 0.63 & 56.32 & 20.27
& 0.40 & 11.58 & 1.10
& 0.63 & 117.53 & 28.98 \\
& Compression Only
& 0.55 & \secondcell{16.74} & \secondcell{2.78}
& 0.38 & 1.91 & 0.29
& 0.59 & \secondcell{9.13} & 2.30 \\
& Schedule Only
& 0.63 & 56.22 & 20.20
& 0.40 & 10.89 & 1.46
& 0.63 & 102.86 & 24.72 \\
& Compress $\rightarrow$ Tune
& 0.53 & 23.67 & 4.19
& 0.39 & \secondcell{1.83} & \secondcell{0.26}
& 0.59 & 10.20 & \secondcell{2.17} \\
& \textbf{CoptV2X (Ours)}
& 0.62 & \bestcell{3.34} & \bestcell{0.65}
& 0.41 & \bestcell{1.50} & \bestcell{0.21}
& 0.60 & \bestcell{8.64} & \bestcell{2.05}\\
\midrule
\multirow{5}{*}{(b)}
& Original/Default
& 0.63 & 3.22 & 1.16
& 0.40 & 1.85 & 0.70
& 0.63 & 11.80 & 6.76 \\
& Compression Only
& 0.54 & \secondcell{0.52} & \secondcell{0.13}
& 0.42 & \secondcell{0.36} & \secondcell{0.10}
& 0.60 & 2.58 & 1.16 \\
& Schedule Only
& 0.63 & 1.10 & 0.43
& 0.40 & 0.77 & 0.35
& 0.63 & 5.15 & 3.49 \\
& Compress $\rightarrow$ Tune
& 0.53 & 1.60 & 0.47
& 0.39 & 0.39 & 0.14
& 0.63 & \secondcell{1.04} & \secondcell{0.57} \\
& \textbf{CoptV2X (Ours)}
& 0.62 & \bestcell{0.45} & \bestcell{0.11}
& 0.40 & \bestcell{0.30} & \bestcell{0.07}
& 0.61 & \bestcell{0.68} & \bestcell{0.28} \\
\midrule
\multirow{5}{*}{(c)}
& Original/Default
& 0.63 & 128.75 & 1.06
& 0.29 & 68.98 & 0.56
& 0.63 & 559.34 & 4.97 \\
& Compression Only
& 0.54 & \secondcell{12.43} & \secondcell{0.11}
& 0.32 & \secondcell{8.48} & \secondcell{0.07}
& 0.60 & \secondcell{41.25} & \secondcell{0.36} \\
& Schedule Only
& 0.63 & 120.94 & 0.83
& 0.21 & 92.49 & 0.71
& 0.63 & 549.63 & 9.58 \\
& Compress $\rightarrow$ Tune
& 0.55 & 36.53 & 0.25
& 0.32 & 10.82 & 0.09
& 0.63 & 48.28 & 0.42 \\
& \textbf{CoptV2X (Ours)}
& 0.62 & \bestcell{10.62} & \bestcell{0.08}
& 0.30 & \bestcell{5.72} & \bestcell{0.05}
& 0.60 & \bestcell{38.24} & \bestcell{0.34} \\
\bottomrule
\end{tabular*}
\caption{Unified deployment comparison under
$\Delta\mathrm{AP}_{70}\leq 0.10$: (a) Default/H800,
(b) TensorRT/H800, and (c) TensorRT/Jetson AGX Orin.
Boldface marks the best and underlining marks the second-best
constraint-feasible latency or energy within each profile and model.}
\label{tab:main-comparison}
\end{table}

Table~\ref{tab:main-comparison}(a) summarizes the results under the Default/H800 profile. For each model, CoptV2X achieves the lowest latency and energy consumption among all feasible candidates. 
Compared with the best feasible decoupled baseline for each model and metric, CoptV2X reduces latency and energy consumption by 34.48\% and 33.79\% on average, respectively, across Pyramid, CoDriving, and F-Cooper. 
Furthermore, with only marginal accuracy degradation, the selected configurations deliver inference speedups of approximately $16.86\times$, $7.72\times$, and $13.60\times$ over the original deployments of the three models, respectively.

The results further indicate that the acceleration gains do not arise solely from reductions in FLOPs. 
\textit{Compression Only} selected FP16 for all three models and obtained no consistent benefit from INT8 execution. 
It also adopted more conservative channel-pruning configurations than CoptV2X. 
For Pyramid, for example, it selected $(24,128,64)$, whereas CoptV2X selected $(16,32,64)$. 
This observation is consistent with the backend profiling results. 
Without scheduling optimization, a single convolutional layer executed in INT8 exhibited 63\% higher latency than the same layer executed in FP16.
This is because lower numerical precision does not guarantee that every operator can use efficient low-precision compute units. Inter-operator precision conversions may also introduce additional latency. 
Likewise, a smaller computation graph does not ensure that the resulting channel dimensions enable efficient TensorRT tactics or satisfy Tensor Core alignment requirements.

\textit{Compress} $\rightarrow$ \textit{Tune} can further optimize operator implementations, memory layouts, and other scheduling decisions after graph compression. 
However, once the computation graph is fixed, its sequential decision process cannot feed backend measurements back into the compression stage. 
Consequently, this workflow faces a limitation similar to that of \textit{Compression Only}. 
The compression stage may prematurely discard channel-width and numerical-precision combinations that would become more efficient after software--hardware co-optimization. 
\textit{Schedule Only} optimizes execution schedules under a fixed computation graph and numerical precision. 
It therefore cannot reduce theoretical computational cost by eliminating redundant computation or lowering numerical precision, limiting its effectiveness for complex networks.

Moreover, CoptV2X's inner-loop optimization is not tied to a specific deployment backend. 
Through a unified interface, it can integrate with efficient compilation backends tailored to the target hardware.
On NVIDIA platforms, CoptV2X can use TensorRT for backend scheduling optimization.
As shown in Table~\ref{tab:main-comparison}(b), CoptV2X reduces latency and energy consumption by 21.58\% and 32.09\% on average, respectively, relative to the best feasible decoupled baseline across the three models.
The framework also supports advanced auto-scheduling systems~\cite{shao2022metaschedule,zheng2020ansor}.

Table~\ref{tab:main-comparison}(c) reports the results on Jetson AGX Orin.
CoptV2X achieves an average inference speedup of $12.94\times$ over the original deployments across three representative V2X perception models.
These gains are consistent with the profiling results in Fig.~\ref{fig:latency-contribution}.
Assuming a transmission latency of 50~$\mathrm{ms}$, Amdahl's law translates these gains into system-level inference speedups of 1.74$\times$ to 2.59$\times$.

Based on the closed-loop simulation results in Fig.~\ref{fig:bg-latency-prune}(a), we further map these results to their corresponding operating scenarios. 
For the compute-intensive F-Cooper model, this mapping indicates that reducing the end-to-end latency from 849~$\mathrm{ms}$ to 328~$\mathrm{ms}$ is expected to increase DS by 49\%.
In contrast, for the computationally lighter Pyramid and CoDriving models, the original end-to-end
latencies already fall within the relatively flat region of the DS--latency curve. 
Although CoptV2X does not yield a substantial DS improvement for these models, it increases the safety margin relative to the latency-collapse threshold. 
This wider margin improves system robustness to communication fluctuations and enables deployment on lower-cost edge computing platforms.

\subsection{Ablation of Online Search Components}

Table~\ref{tab:online-search-ablation} evaluates how the core components of the bilevel search architecture affect search efficiency under a limited measurement budget.
We evaluate search efficiency using $\Delta$HV-AUC, defined in Equations~\ref{eq:delta-hv} and~\ref{eq:delta-hv-auc}, which measures Pareto-front expansion over the entire search trajectory:

\begin{equation}
\Delta\mathrm{HV}_{\tau}
=\left[\mathrm{HV}\!\left(\mathcal{D}_{\mathrm{out}}^{\tau}\right)
-\mathrm{HV}\!\left(\mathcal{D}_{\mathrm{out}}^{0}\right)\right]_+.
\label{eq:delta-hv}
\end{equation}
\begin{equation}
\Delta\mathrm{HV}\text{-}\mathrm{AUC}
=2\sum_{\tau=1}^{4}
\left(\Delta\mathrm{HV}_{\tau-1}+\Delta\mathrm{HV}_{\tau}\right).
\label{eq:delta-hv-auc}
\end{equation}
where $\mathcal{D}_{\mathrm{out}}^{0}$ and $\mathcal{D}_{\mathrm{out}}^{\tau}$ denote the cold-start data and measurements accumulated through round $\tau$, respectively;
$\mathrm{HV}(\cdot)$ denotes the hypervolume computed against the fixed reference point in the normalized $(\mathrm{latency},\mathrm{energy},-\mathrm{AP}_{70})$ objective space;
and $[\cdot]_+$ retains the non-negative part of the enclosed difference.
Each round selects four candidates, yielding budget checkpoints of $(0,4,8,12,16)$ and a trapezoidal-integration coefficient of 2.
Thus, $\Delta\mathrm{HV}\text{-}\mathrm{AUC}$ captures both the extent and timing of Pareto-front expansion.
A larger value indicates that high-quality Pareto candidates are found earlier and expand the frontier further.

Removing graph features produced the largest decrease in $\Delta\mathrm{HV}\text{-}\mathrm{AUC}$, at 63.1\%. 
Without deployment-relevant graph information, the cost models could not adequately distinguish how different configurations execute on the target hardware, making it more difficult for the outer loop to identify promising candidates. 
Replacing cost-model-guided selection with uniform random sampling reduced the metric by 34.0\%, because the limited measurement budget was no longer allocated according to predicted multi-metric trade-offs.
Freezing online measured feedback reduced it by 46.9\%, because the cost models could no longer use newly acquired hardware measurements and deployment evidence to correct cold-start prediction errors.

\begin{table}[t]
\centering
\setlength{\tabcolsep}{2.5pt}
\begin{tabular*}{\columnwidth}{@{\extracolsep{\fill}}lcccrr@{}}
\toprule
& \multicolumn{3}{c}{Components} & \multicolumn{2}{c}{Search quality} \\
\cmidrule(lr){2-4}\cmidrule(lr){5-6}
Variant & CM & FB & GF & $\Delta$HV-AUC$\uparrow$ & $\Delta$ \\
\midrule
\textbf{CoptV2X} & $\checkmark$ & $\checkmark$ & $\checkmark$ & \textbf{15.90$\pm$4.47} & -- \\
w/o cost model & -- & $\checkmark$ & $\checkmark$ & 10.50$\pm$6.14 & $-5.40$ \\
w/o feedback & $\checkmark$ & -- & $\checkmark$ & 8.44$\pm$0.62 & $-7.46$ \\
w/o graph features & $\checkmark$ & $\checkmark$ & -- & 5.87$\pm$2.51 & $-10.03$ \\
\bottomrule
\end{tabular*}
\caption{Ablation of online search components (mean $\pm$ s.d. over three seeds). CM, FB, and GF denote cost-model guidance, online measured feedback, and graph features, respectively. $\Delta$ denotes the change with respect to CoptV2X.}
\label{tab:online-search-ablation}
\end{table}

\section{Conclusion}

In this work, we presented CoptV2X, a software--hardware co-optimization framework for efficient V2X perception deployment. 
CoptV2X constructs a hardware-aware search space by combining backend capability profiling with computation-graph dependency partitioning. 
Its bilevel search uses hardware feedback to coordinate outer-loop pruning and quantization with inner-loop schedule tuning, while multi-metric cost models reduce measurement overhead. 
Across multiple representative V2X models and backends, CoptV2X reduces latency and energy by 28.03\% and 32.94\% on average over the strongest feasible decoupled baseline on H800 under $\Delta\mathrm{AP}_{70}\leq0.10$, while achieving an average inference speedup of $12.94\times$ over the original deployments on Jetson AGX Orin.
Overall, this study further substantiates the tight coupling between model compression and backend execution optimization and offers a new perspective on the efficient deployment of V2X perception models.

\FloatBarrier
\bibliographystyle{unsrtnat}
\bibliography{coptv2x}

@inproceedings{balasubramaniam2025upaq,
  title={{UPAQ}: A Framework for Real-Time and Energy-Efficient 3D Object Detection in Autonomous Vehicles},
  author={Balasubramaniam, Abhishek and Sunny, Febin P. and Pasricha, Sudeep},
  booktitle={Proceedings of the 2025 Design, Automation \& Test in Europe Conference (DATE)},
  pages={1--7},
  year={2025},
  doi={10.23919/DATE64628.2025.10993184}
}

@inproceedings{chen2016xgboost,
  author    = {Chen, Tianqi and Guestrin, Carlos},
  title     = {{XGBoost}: A Scalable Tree Boosting System},
  booktitle = {Proceedings of the 22nd ACM SIGKDD International
               Conference on Knowledge Discovery and Data Mining},
  pages     = {785--794},
  year      = {2016},
  doi       = {10.1145/2939672.2939785}
}

@inproceedings{chen2018learning,
  title={Learning to Optimize Tensor Programs},
  author={Chen, Tianqi and Zheng, Lianmin and Yan, Eddie and Jiang, Ziheng and Moreau, Thierry and Ceze, Luis and Guestrin, Carlos and Krishnamurthy, Arvind},
  booktitle={Advances in Neural Information Processing Systems (NeurIPS)},
  volume={31},
  pages = {3393--3404},
  year={2018}
}

@inproceedings{chen2019fcooper,
  title     = {{F-Cooper}: Feature-Based Cooperative Perception
               for Autonomous Vehicle Edge Computing System
               Using 3D Point Clouds},
  author    = {Chen, Qi and Ma, Xu and Tang, Sihai and
               Guo, Jingda and Yang, Qing and Fu, Song},
  booktitle = {Proceedings of the 4th ACM/IEEE Symposium
               on Edge Computing},
  pages     = {88--100},
  year      = {2019},
  doi       = {10.1145/3318216.3363300}
}

@inproceedings{chen2021chanas,
  title={{CHaNAS}: Coordinated Search for Network Architecture and Scheduling Policy},
  author={Chen, Weiwei and Wang, Ying and Lin, Gangliang and Gao, Chengsi and Liu, Cheng and Zhang, Lei},
  booktitle={Proceedings of the 22nd ACM SIGPLAN/SIGBED International Conference on Languages, Compilers, and Tools for Embedded Systems (LCTES)},
  pages={42--53},
  year={2021},
  doi={10.1145/3461648.3463846}
}

@article{deb2002nsga,
  title={A Fast and Elitist Multiobjective Genetic Algorithm: {NSGA-II}},
  author={Deb, Kalyanmoy and Pratap, Amrit and Agarwal, Sameer and Meyarivan, T},
  journal={IEEE Transactions on Evolutionary Computation},
  volume={6},
  number={2},
  pages={182--197},
  year={2002}
}

@inproceedings{fang2023depgraph,
  title={DepGraph: Towards Any Structural Pruning},
  author={Fang, Gongfan and Ma, Xinyin and Song, Mingli and Mi, Michael Bi and Wang, Xinchao},
  booktitle={Proceedings of the IEEE/CVF Conference on Computer Vision and Pattern Recognition (CVPR)},
  pages={16091--16101},
  year={2023}
}

@article{geurts2006extremely,
  author  = {Geurts, Pierre and Ernst, Damien and Wehenkel, Louis},
  title   = {Extremely Randomized Trees},
  journal = {Machine Learning},
  volume  = {63},
  number  = {1},
  pages   = {3--42},
  year    = {2006},
  doi     = {10.1007/s10994-006-6226-1}
}

@inproceedings{grinsztajn2022tree,
  title     = {Why Do Tree-Based Models Still Outperform Deep Learning on Typical Tabular Data?},
  author    = {Grinsztajn, Leo and Oyallon, Edouard and Varoquaux, Gael},
  booktitle = {Advances in Neural Information Processing Systems},
  volume    = {35},
  pages     = {507--520},
  year      = {2022}
}

@inproceedings{hu2021opq,
  title={{OPQ}: Compressing Deep Neural Networks with One-shot Pruning-Quantization},
  author={Hu, Peng and Peng, Xi and Zhu, Hongyuan and Aly, Mohamed M. Sabry and Lin, Jie},
  booktitle={Proceedings of the AAAI Conference on Artificial Intelligence},
  volume={35},
  pages={7780--7788},
  year={2021},
  doi={10.1609/aaai.v35i9.16950}
}

@inproceedings{ke2017lightgbm,
  author    = {Ke, Guolin and Meng, Qi and Finley, Thomas and
               Wang, Taifeng and Chen, Wei and Ma, Weidong and
               Ye, Qiwei and Liu, Tie-Yan},
  title     = {{LightGBM}: A Highly Efficient Gradient Boosting
               Decision Tree},
  booktitle = {Advances in Neural Information Processing Systems},
  volume    = {30},
  pages     = {3146--3154},
  year      = {2017}
}

@inproceedings{lee2023flexround,
  title     = {{FlexRound}: Learnable Rounding Based on Element-Wise Division
               for Post-Training Quantization},
  author    = {Lee, Jung Hyun and Kim, Jeonghoon and Kwon, Se Jung and Lee, Dongsoo},
  booktitle = {Proceedings of the 40th International Conference on Machine Learning},
  volume    = {202},
  pages     = {18913--18939},
  series    = {Proceedings of Machine Learning Research},
  year      = {2023}
}

@article{lindauer2022smac3,
  title={{SMAC3}: A Versatile Bayesian Optimization Package for Hyperparameter Optimization},
  author={Lindauer, Marius and Eggensperger, Katharina and Feurer, Matthias and Biedenkapp, Andr{\'e} and Deng, Difan and Benjamins, Carolin and Ruhkopf, Tim and Sass, Ren{\'e} and Hutter, Frank},
  journal={Journal of Machine Learning Research},
  volume={23},
  number={54},
  pages={1--9},
  year={2022}
}

@article{liu2025toward,
  title={Toward Collaborative Autonomous Driving: Simulation Platform and End-to-End System},
  author={Liu, Genjia and Hu, Yue and Xu, Chenxin and Mao, Weibo and Ge, Junhao and Huang, Zhengxiang and Lu, Yifan and Xu, Yinda and Xia, Junkai and Wang, Yafei and Chen, Siheng},
  journal={IEEE Transactions on Pattern Analysis and Machine Intelligence},
  volume={47},
  number={8},
  pages={6566--6584},
  year={2025},
  doi={10.1109/TPAMI.2025.3560327}
}

@inproceedings{liu2026ipda,
  title={{IPDA}: Intelligent Perception Delay Alignment Method Based on Spatio-Temporal Co-Sensing Calibration},
  author={Liu, Jianhang and Zhang, Dianzheng and Pan, Hongxin and Jiang, Guangqian and Fan, Runda},
  booktitle={Proceedings of the AAAI Conference on Artificial Intelligence},
  volume={40},
  pages={29546--29554},
  year={2026},
  doi={10.1609/aaai.v40i35.40197}
}

@inproceedings{lu2024heal,
  title     = {An Extensible Framework for Open Heterogeneous
               Collaborative Perception},
  author    = {Lu, Yifan and Hu, Yue and Zhong, Yiqi and
               Wang, Dequan and Wang, Yanfeng and Chen, Siheng},
  booktitle = {International Conference on Learning Representations},
  year      = {2024}
}

@article{rook2026mosmac,
  title   = {{MO-SMAC}: Multi-objective Sequential Model-based
             Algorithm Configuration},
  author  = {Rook, Jeroen G. and Benjamins, Carolin and
             Bossek, Jakob and Trautmann, Heike and
             Hoos, Holger H. and Lindauer, Marius},
  journal = {Evolutionary Computation},
  volume  = {34},
  number  = {1},
  pages   = {29--52},
  year    = {2026},
  doi     = {10.1162/evco_a_00371}
}

@inproceedings{shao2022metaschedule,
  title={Tensor Program Optimization with Probabilistic Programs},
  author={Shao, Junru and Zhou, Xiyou and Feng, Siyuan and Hou, Bohan and Lai, Ruihang and Jin, Hongyi and Lin, Wuwei and Masuda, Masahiro and Yu, Cody Hao and Chen, Tianqi},
  booktitle={Advances in Neural Information Processing Systems (NeurIPS)},
  volume={35},
  pages={35783--35796},
  year={2022}
}

@inproceedings{shen2022halp,
  title={Structural Pruning via Latency-Saliency Knapsack},
  author={Shen, Maying and Yin, Hongxu and Molchanov, Pavlo and Mao, Lei and Liu, Jianna and Alvarez, Jose M.},
  booktitle={Advances in Neural Information Processing Systems (NeurIPS)},
  volume={35},
  pages={12894--12908},
  year={2022}
}

@inproceedings{tang2020semi,
  title     = {A Semi-Supervised Assessor of Neural Architectures},
  author    = {Tang, Yehui and Wang, Yunhe and Xu, Yixing and Chen, Hanting
               and Shi, Boxin and Xu, Chao and Xu, Chunjing and Tian, Qi
               and Xu, Chang},
  booktitle = {Proceedings of the IEEE/CVF Conference on Computer Vision
               and Pattern Recognition},
  pages     = {1810--1819},
  year      = {2020}
}

@inproceedings{wang2019haq,
  title={{HAQ}: Hardware-Aware Automated Quantization with Mixed Precision},
  author={Wang, Kuan and Liu, Zhijian and Lin, Yujun and Lin, Ji and Han, Song},
  booktitle={Proceedings of the IEEE/CVF Conference on Computer Vision and Pattern Recognition (CVPR)},
  pages={8612--8620},
  year={2019}
}

@inproceedings{wang2026cpfreezer,
  title={{CP-FREEZER}: Latency Attacks Against Vehicular Cooperative Perception},
  author={Wang, Chenyi and Song, Ruoyu and Muller, Raymond and Monteuuis, Jean-Philippe and Celik, Z. Berkay and Petit, Jonathan and Gerdes, Ryan and Li, Ming},
  booktitle={Proceedings of the AAAI Conference on Artificial Intelligence},
  volume={40},
  pages={1114--1122},
  year={2026},
  doi={10.1609/aaai.v40i2.37082}
}

@inproceedings{wang2026sparsecoop,
  title={{SparseCoop}: Cooperative Perception with Kinematic-Grounded Queries},
  author={Wang, Jiahao and Jiang, Zhongwei and Sun, Wenchao and Zhong, Jiaru and Yu, Haibao and Zhang, Yuner and Lu, Chenyang and Zhang, Chuang and He, Lei and Xu, Shaobing and Wang, Jianqiang},
  booktitle={Proceedings of the AAAI Conference on Artificial Intelligence},
  volume={40},
  pages={9876--9884},
  year={2026},
  doi={10.1609/aaai.v40i12.37952}
}

@article{williams2009roofline,
  title={Roofline: An Insightful Visual Performance Model for Multicore Architectures},
  author={Williams, Samuel and Waterman, Andrew and Patterson, David},
  journal={Communications of the ACM},
  volume={52},
  number={4},
  pages={65--76},
  year={2009},
  doi={10.1145/1498765.1498785}
}

@inproceedings{xiao2023haloc,
  title={{HALOC}: Hardware-Aware Automatic Low-Rank Compression for Compact Neural Networks},
  author={Xiao, Jinqi and Zhang, Chengming and Gong, Yu and Yin, Miao and Sui, Yang and Xiang, Lizhi and Tao, Dingwen and Yuan, Bo},
  booktitle={Proceedings of the AAAI Conference on Artificial Intelligence},
  volume={37},
  pages={10464--10472},
  year={2023},
  doi={10.1609/aaai.v37i9.26244}
}

@inproceedings{xu2022opv2v,
  title     = {{OPV2V}: An Open Benchmark Dataset and Fusion
               Pipeline for Perception with Vehicle-to-Vehicle
               Communication},
  author    = {Xu, Runsheng and Xiang, Hao and Xia, Xin and
               Han, Xu and Li, Jinlong and Ma, Jiaqi},
  booktitle = {2022 IEEE International Conference on
               Robotics and Automation},
  pages     = {2583--2589},
  year      = {2022},
  doi       = {10.1109/ICRA46639.2022.9812038}
}

@inproceedings{xu2023alt,
  title     = {{ALT}: Breaking the Wall between Data Layout and
               Loop Optimizations for Deep Learning Compilation},
  author    = {Xu, Zhiying and Xu, Jiafan and Peng, Hongding and
               Wang, Wei and Wang, Xiaoliang and Wan, Haoran and
               Dai, Haipeng and Xu, Yixu and Cheng, Hao and
               Wang, Kun and Chen, Guihai},
  booktitle = {Proceedings of the Eighteenth European Conference
               on Computer Systems},
  pages     = {199--214},
  year      = {2023},
  doi       = {10.1145/3552326.3587440}
}

@inproceedings{yang2021fracbits,
  title={{FracBits}: Mixed Precision Quantization via Fractional Bit-Widths},
  author={Yang, Linjie and Jin, Qing},
  booktitle={Proceedings of the AAAI Conference on Artificial Intelligence},
  volume={35},
  pages={10612--10620},
  year={2021},
  doi={10.1609/aaai.v35i12.17269}
}

@inproceedings{yu2022dairv2x,
  title     = {{DAIR-V2X}: A Large-Scale Dataset for
               Vehicle-Infrastructure Cooperative 3D Object Detection},
  author    = {Yu, Haibao and Luo, Yizhen and Shu, Mao and
               Huo, Yiyi and Yang, Zebang and Shi, Yifeng and
               Guo, Zhenglong and Li, Hanyu and Hu, Xing and
               Yuan, Jirui and Nie, Zaiqing},
  booktitle = {Proceedings of the IEEE/CVF Conference on
               Computer Vision and Pattern Recognition},
  pages     = {21361--21370},
  year      = {2022}
}

@inproceedings{yu2025univ2x,
  title={End-to-End Autonomous Driving Through {V2X} Cooperation},
  author={Yu, Haibao and Yang, Wenxian and Zhong, Jiaru and Yang, Zhenwei and Fan, Siqi and Luo, Ping and Nie, Zaiqing},
  booktitle={Proceedings of the AAAI Conference on Artificial Intelligence},
  volume={39},
  pages={9598--9606},
  year={2025},
  doi={10.1609/aaai.v39i9.33040}
}

@inproceedings{zhao2019variational,
  title     = {Variational Convolutional Neural Network Pruning},
  author    = {Zhao, Chenglong and Ni, Bingbing and Zhang, Jian and Zhao, Qiwei
               and Zhang, Wenjun and Tian, Qi},
  booktitle = {Proceedings of the IEEE/CVF Conference on Computer Vision
               and Pattern Recognition},
  pages     = {2780--2789},
  year      = {2019}
}

@article{zhao2025quantv2x,
  title         = {{QuantV2X}: A Fully Quantized Multi-Agent System for Cooperative Perception},
  author        = {Zhao, Seth Z. and Zhang, Huizhi and Li, Zhaowei and Peng, Juntong and Chui, Anthony and Zhou, Zewei and Meng, Zonglin and Xiang, Hao and Huang, Zhiyu and Wang, Fujia and Tian, Ran and Xu, Chenfeng and Zhou, Bolei and Ma, Jiaqi},
  journal       = {arXiv preprint arXiv:2509.03704},
  year          = {2025},
  eprint        = {2509.03704},
  archivePrefix = {arXiv},
  primaryClass  = {cs.CV},
  doi           = {10.48550/arXiv.2509.03704}
}

@inproceedings{zheng2020ansor,
  title     = {{Ansor}: Generating High-Performance Tensor Programs for Deep Learning},
  author    = {Zheng, Lianmin and Jia, Chengfan and Sun, Minmin and Wu, Zhao and Yu, Cody Hao and Haj-Ali, Ameer and Wang, Yida and Yang, Jun and Zhuo, Danyang and Sen, Koushik and Gonzalez, Joseph E. and Stoica, Ion},
  booktitle = {14th USENIX Symposium on Operating Systems Design and Implementation (OSDI)},
  pages     = {863--879},
  year      = {2020},
  publisher = {USENIX Association},
  isbn      = {978-1-939133-19-9}
}

@inproceedings{zheng2022dietcode,
  title={{DietCode}: Automatic Optimization for Dynamic Tensor Programs},
  author={Zheng, Bojian and Jiang, Ziheng and Yu, Cody Hao and Shen, Haichen and Fromm, Joshua and Liu, Yizhi and Wang, Yida and Ceze, Luis and Chen, Tianqi and Pekhimenko, Gennady},
  booktitle={Proceedings of Machine Learning and Systems (MLSys)},
  volume={4},
  pages={848--863},
  year={2022}
}

@inproceedings{zhu2022roller,
  title={{ROLLER}: Fast and Efficient Tensor Compilation for Deep Learning},
  author={Zhu, Hongyu and Wu, Ruofan and Diao, Yijia and Ke, Shanbin and Li, Haoyu and Zhang, Chen and Xue, Jilong and Ma, Lingxiao and Xia, Yuqing and Cui, Wei and Yang, Fan and Yang, Mao and Zhou, Lidong and Cidon, Asaf and Pekhimenko, Gennady},
  booktitle={16th USENIX Symposium on Operating Systems Design and Implementation (OSDI)},
  pages={233--248},
  year={2022}
}

\clearpage
\appendix
\section{Cost-Model and Target-Transformation Selection}
\label{app:cost-model-selection}

\begin{figure}[!htbp]
\centering
\includegraphics[width=\columnwidth]{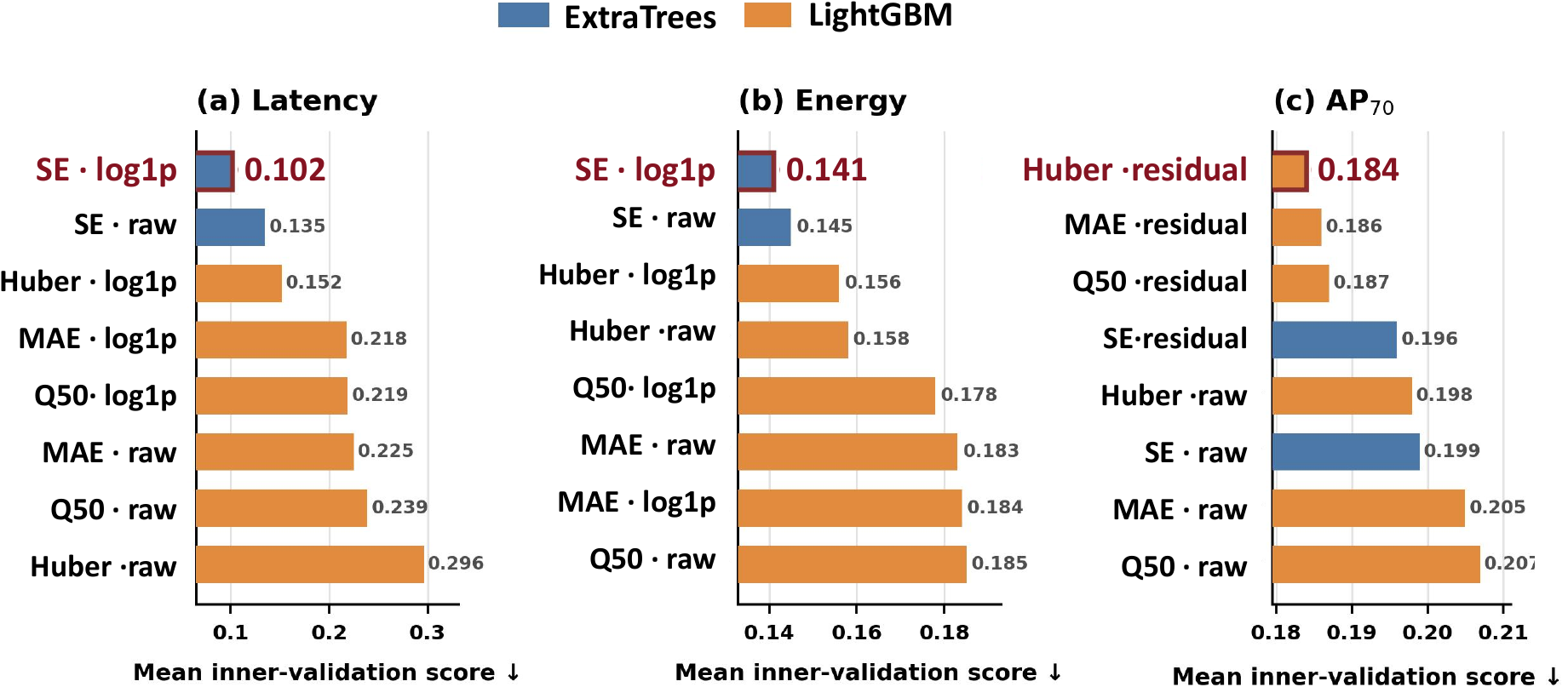}
\caption{Cost-model and target-transformation selection under nested grouped cross-validation. Lower scores are better but not comparable across metrics.}
\label{fig:cost-model-selection}
\end{figure}

The cost-model class and target transformation are selected by nested grouped cross-validation with five outer folds and three inner folds. 
The analysis uses 176 real measurements covering 44 complete model--structure groups, with all samples from a group kept together in every split.
Within each outer fold, predictor--transformation pairs are ranked by their mean score across the three inner folds. The top-ranked pair is then refitted on the complete outer-training set and evaluated on the held-out outer fold.
The resulting selection is shown in Fig.~\ref{fig:cost-model-selection}. 
The horizontal axis reports the mean inner-validation score, aggregated across the five outer splits, as defined in Equation~\ref{eq:mean-inner-score}:

\begin{equation}
\mathrm{MIS}(m,t)=\frac{1}{5}\sum_{u=1}^{5}\frac{1}{3}\sum_{v=1}^{3}S_{u,v}(m,t),
\label{eq:mean-inner-score}
\end{equation}
where $(m,t)$ denotes a predictor and target-transformation pair and the fold-level score $S_{u,v}$ is defined in Equation~\ref{eq:inner-score}:

\begin{equation}
S_{u,v}=\frac{\mathrm{MAE}_{u,v}}{\max(\Delta_{u,v},10^{-9})}+0.25(1-\mathrm{SRCC}_{u,v}),
\label{eq:inner-score}
\end{equation}
where $\mathrm{MAE}_{u,v}$ is the mean absolute error on the inner fold, $\Delta_{u,v}=Q_{0.9}(y_{u,v})-Q_{0.1}(y_{u,v})$ is the robust label span of inner fold $(u,v)$, and $\mathrm{SRCC}_{u,v}$ is the Spearman rank correlation coefficient. 
The first term measures normalized prediction error and the second penalizes incorrect candidate ordering; hence, lower scores indicate better predictive performance.

Based on this criterion, ExtraTrees regressors with a \texttt{log1p}-transformed target are selected for both latency and energy, whereas a LightGBM predictor trained with the Huber loss and a model-anchored residual target is selected for $\mathrm{AP}_{70}$.

\end{document}